\documentclass{article} 
\usepackage{iclr2027_conference,times}

\usepackage{amsmath,amsfonts,bm}

\def\eqref#1{equation~\ref{#1}}

\def\1{\bm{1}}

\DeclareMathAlphabet{\mathsfit}{\encodingdefault}{\sfdefault}{m}{sl}
\SetMathAlphabet{\mathsfit}{bold}{\encodingdefault}{\sfdefault}{bx}{n}

\usepackage{url}
\usepackage{lipsum} 
\usepackage[dvipsnames]{xcolor}

\usepackage{booktabs}
\usepackage{graphicx}
\usepackage{amsmath}
\usepackage{amssymb}
\usepackage{xcolor}
\usepackage{float}
\usepackage{capt-of}
\usepackage{listings}
\usepackage{xspace}
\usepackage{tikz}
\usepackage{amssymb}
\usepackage{wrapfig}

\usepackage{pifont}
\DeclareRobustCommand{\cmark}{\mbox{\textcolor{ForestGreen}{\ding{51}}}}
\DeclareRobustCommand{\xmark}{\mbox{\textcolor{black!35}{\ding{55}}}}
\DeclareRobustCommand{\na}{\textcolor{black!35}{--}}
\usetikzlibrary{arrows.meta,positioning,fit,backgrounds,calc}
\usepackage{fontawesome5}
\definecolor{slate}{HTML}{3A4A6B}
\definecolor{aipurple}{HTML}{7B4BC0}
\definecolor{judge}{HTML}{2B6CB0}
\definecolor{okgreen}{HTML}{2F7D32}
\definecolor{failred}{HTML}{B03030}

\definecolor{citecolor}{HTML}{0071BC}
\definecolor{linkcolor}{HTML}{ED1C24}
\usepackage[pagebackref=true, breaklinks=true, colorlinks,
            citecolor=citecolor, linkcolor=linkcolor, bookmarks=false]{hyperref}

\newcommand{\bench}{\textsc{EvoWeb}\xspace} 

\title{EvoWeb: A Benchmark for Coding Agents on Evolving Web Applications}

\author{%
David Nukrai$^{\Delta*}$\quad Bar Alon$^{\Delta,\text{\scalebox{1.35}{$\boldsymbol{\tau}$}}*}$\quad Nativ Levy$^{\Delta}$\quad Ido Andrew Atad$^{\Delta}$\quad Ido Cohen$^{\Delta}$\\[2pt]
Itamar Zimerman$^{\Delta}$\quad Lior Wolf$^{\text{\scalebox{1.35}{$\boldsymbol{\tau}$}}}$\quad Eli David$^{\Delta}$\quad Ilan Naiman$^{\Delta}$\\[4pt]
{\normalfont\small $^{\Delta}$Stealth Startup, Tel Aviv\qquad $^{\text{\scalebox{1.35}{$\boldsymbol{\tau}$}}}$Tel Aviv University\qquad $^{*}$Equal contribution}%
}

\makeatletter
\def\@maketitle{\vbox{\hsize\textwidth\centering
{\LARGE\sc \@title\par}\vskip 0.17in
{\bf \@author\par}\vskip 0.12in}}
\makeatother
\definecolor{hfyellow}{HTML}{FFD21E}\definecolor{hforange}{HTML}{FF9D0B}\definecolor{hfdark}{HTML}{3A3B45}
\newcommand{\hficon}{\raisebox{-0.35ex}{\tikz[x=0.95ex,y=0.95ex]{\fill[hfyellow] (0,0) circle (1);\fill[hfdark] (-0.36,0.28) circle (0.13);\fill[hfdark] (0.36,0.28) circle (0.13);\fill[hfdark] (-0.42,-0.12) arc (180:360:0.42) -- cycle;\fill[hforange,rotate around={35:(-0.82,-0.55)}] (-0.82,-0.55) ellipse (0.32 and 0.22);\fill[hforange,rotate around={-35:(0.82,-0.55)}] (0.82,-0.55) ellipse (0.32 and 0.22);}}}
\newsavebox{\hfbox}
\AtBeginDocument{\sbox{\hfbox}{\hficon}}
\usetikzlibrary{shadows}
\newcommand{\linkbadge}[2]{
\href{#1}{\tikz[baseline=(b.base)]\node[rounded corners=5pt,fill=black!2.5,draw=black!18,line width=0.45pt,text=black!82,minimum width=2.45cm,inner ysep=3.2pt,font=\sffamily\small,drop shadow={opacity=0.12,shadow xshift=0.35pt,shadow yshift=-0.7pt}](b){\strut #2};}}
\hypersetup{pdftitle={EvoWeb: A Benchmark for Coding Agents on Evolving Web Applications},pdfauthor={David Nukrai, Bar Alon, Nativ Levy, Ido Andrew Atad, Ido Cohen, Itamar Zimerman, Lior Wolf, Eli David, Ilan Naiman}}

\iclrfinalcopy 
\begin{document}

\maketitle
\vspace{-7pt}
\begin{abstract}
Web application development often proceeds through successive requests that extend an existing implementation. Yet web-generation benchmarks primarily evaluate development toward fixed specifications. We introduce EvoWeb, a novel benchmark comprising 320 cumulative requests over 32 web applications, grounded in real-world product features and structured by functional dependency graphs. Agents iteratively modify a persistent codebase, while a multimodal web agent tests each intermediate application in a live browser, assessing newly requested functionality, cross-request integration, and previously requested functionality. To characterize reliability throughout development, we integrate trajectory performance across user-tolerance thresholds rather than impose a fixed stopping criterion. Experiments across open- and closed-source frontier models show that EvoWeb reveals clear performance differences between models, with cross-request integration emerging as a key challenge. Further analysis shows that later requests with deeper or more numerous dependencies are solved less accurately, while runtime errors become more frequent over successive requests. EvoWeb provides a testbed for evaluating and improving coding agents in long-horizon, multi-request development.

\newlength{\githubiconheight}
\settoheight{\githubiconheight}{{\large\faGithub}}

\vspace{-1pt}
{\centering\hypersetup{pdfborder={0 0 0}}%
\linkbadge{https://github.com/two-delta/EvoWeb}{{\large\faGithub}\hspace{5pt}Code}\hspace{12pt}%
\linkbadge{https://huggingface.co/datasets/TD-Labs/EvoWeb}{%
\raisebox{0.15ex}{{\resizebox{!}{\githubiconheight}{\usebox{\hfbox}}}}%
\hspace{5pt}Dataset}\par}
\end{abstract}

\vspace{-6pt}
\section{Introduction}
\vspace{-2pt}
Large language models are increasingly used as agents for end-to-end software development, where they are responsible not only for generating code but for translating high-level user requests into complete, working applications. Systems such as Bolt.new and Lovable make this capability accessible to users with limited or no programming expertise through natural-language interaction. Users describe the functionality and appearance they want, while the agent decides how to implement them and is responsible for producing a working application~\citep{lu2025webgenbench}. Development in this setting often
continues beyond the initial request. Users return to add features, modify
existing behavior, and extend functionality introduced earlier. This changes what it means to evaluate a coding agent. A later request may
depend on state, interfaces, or behavior that the same agent implemented several
requests earlier. A new feature may work on its own but fail when it needs to
use that earlier functionality. It may also break behavior that previously
worked. Evaluating requests independently therefore does not capture whether an
agent can continue extending its own application while making functionality
introduced at different points work together.

Existing web-generation benchmarks evaluate either applications produced from a single high-level request~\citep{lu2025webgenbench,liu2026webcoderbench,tran2026vibecodebench}
or multiple interactions that refine an application toward an existing
specification or user intent~\citep{lu2025webgenagent,wang2026asukabench}. Persistent multi-request
development has also been studied in software engineering~\citep{evocodebench2026,wu2026swetogether,sobal2026staminabenchstresstestingcodingagents,2026arXiv260324755O}. None of these targets user-facing web applications that evolve through successive requests. This leaves a central question:\par
{\centering\emph{How reliably can coding agents extend a persistent application over successive user requests while integrating with previously implemented functionality and preserving existing behavior?}\par}

We introduce \bench, a benchmark for evaluating coding agents on evolving web
applications. \bench contains 320 cumulative requests over 32 applications. Starting from an initial application specification,
we ground follow-up requests in functionality observed in real-world web
applications, so that applications grow through plausible product features.
We jointly construct each request sequence, its verification tests, and a
directed acyclic graph (DAG) of functional dependencies, where an edge indicates that a later request requires functionality introduced by an earlier one. The resulting sequences undergo automated verification and repair followed by
human curation. The DAG is hidden from the agent, which receives requests sequentially and
modifies the application produced by all preceding requests. For each follow-up request, we define five browser-based verification tests: three test the newly requested functionality, one exercises its integration with earlier
functionality along a dependency in the DAG, and one re-tests functionality requested earlier. After every request, these tests are executed against the resulting application in the browser. This allows us to distinguish whether the agent can implement new functionality, deliver earlier-requested functionality in the current application, and correctly compose functionality introduced across
different points in the development trajectory.

\begin{figure}[t]
\centering
\includegraphics[width=0.99\linewidth]{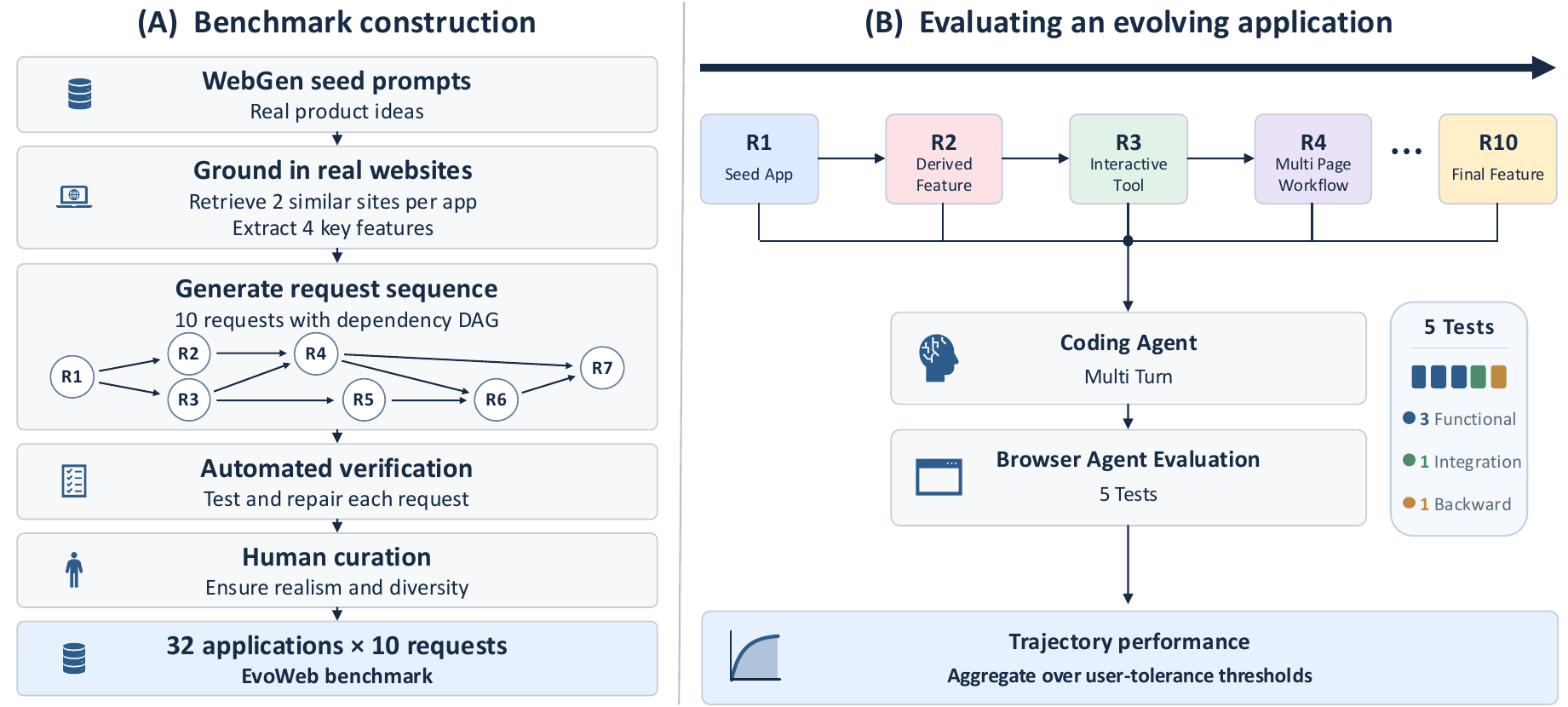}
\vspace{-8pt}
\caption{\textbf{\bench{} pipeline}. \textbf{(A)} Construction: for each seed prompt, features from two similar real websites are used to plan a dependency DAG and write ten requests, which are automatically verified, repaired, and human-curated. \textbf{(B)} Evaluation: a coding agent solves them one at a time, a browser agent runs five tests after each, and scores are averaged over user-tolerance thresholds.}
\label{fig:pipeline}
\vspace{-13pt}
\end{figure}

Sequential evaluation also requires deciding what happens after an intermediate
failure. Stopping at the first failure assumes that the user abandons
development immediately, while continuing regardless of quality assumes that
the user always continues. More generally, the measured trajectory can depend
on the success threshold used to decide whether development continues. We
therefore evaluate each trajectory across a range of user-tolerance thresholds
and integrate performance over these thresholds. This avoids committing to a single stopping rule.

Our evaluation of 14 coding models on \bench shows that reliably extending an application across successive requests remains challenging. Fable 5 and Kimi K3 lead with Grades of 0.694 and 0.684. Cross-request integration consistently scores lower than new or earlier-requested functionality, highlighting the difficulty of making features introduced at different stages work together. Among later requests, accuracy also declines with dependency depth and the number of earlier requests they depend on. Additionally, we observe runtime errors become more frequent over successive requests, while build failures remain rare for most models. These performance differences are not simply a matter of generation effort: Fable 5 scores at the top while using the fewest agent turns and less than half the output tokens of Sonnet 4.6. Our main contributions can be summarized as follows:

\textbf{Benchmark.} We introduce \bench, a benchmark of 320 cumulative requests over 32 web applications, for evaluating coding agents as applications evolve.

\textbf{Methodology.} We develop a methodology for constructing and evaluating multi-request development trajectories, including DAG-structured dependencies across requests and tolerance-averaged metrics that capture trajectory performance.

\textbf{Empirical study.} We provide a broad empirical study of coding agents on evolving web applications and find that cross-request integration is consistently harder than implementing new functionality or delivering earlier-requested functionality, revealing a key source of failure as applications grow across requests.

\begin{table}[t]
\centering
\caption{Comparison with related benchmarks.
Fail-stop and Independent scoring are the $t{=}1$ and $t{=}0$ extremes of our delivered functionality $Q(t)$ (Sec.~\ref{sec:metrics}).}
\label{tab:benchmark-comparison}
\scriptsize
\setlength{\tabcolsep}{2.5pt}
\renewcommand{\arraystretch}{1.1}
\begin{tabular*}{\linewidth}{@{\extracolsep{\fill}}lcccccc c@{}}
\toprule
& \multicolumn{2}{c}{Setting} & \multicolumn{3}{c}{Interaction} & \multicolumn{2}{c}{Scoring} \\
\cmidrule(lr){2-3}\cmidrule(lr){4-6}\cmidrule(l){7-8}
Benchmark & Web app & Request type & Multi-turn & Multi-request & Dependencies & Request & Trajectory \\
\midrule
SlopCodeBench~(\citeauthor{2026arXiv260324755O})& \xmark & SWE     & \cmark & \cmark & Sequential & Binary & Independent \\
EvoCode-Bench~(\citeauthor{evocodebench2026})   & \xmark & SWE     & \cmark & \cmark & Sequential & Binary & Fail-stop \\
StaminaBench~(\citeauthor{sobal2026staminabenchstresstestingcodingagents})
                                         & \xmark & SWE     & \cmark & \cmark & Sequential & Binary & Fail-stop \\
\midrule
WebGen-Bench~(\citeauthor{lu2025webgenbench})   & \cmark & Product & \xmark & \xmark & \na        & Graded & \na \\
WebCoderBench~(\citeauthor{liu2026webcoderbench}) & \cmark & Product & \xmark & \xmark & \na  & Graded & \na \\
Vibe Code Bench~(\citeauthor{tran2026vibecodebench}) & \cmark & Product & \xmark & \xmark & \na & Graded & \na \\
Asuka-Bench~(\citeauthor{wang2026asukabench})   & \cmark & Product & \cmark & \xmark & \na        & Graded & Final \\
Web-Bench~(\citeauthor{xu2025webbench})         & \cmark & SWE     & \cmark & \cmark & Sequential & Binary & Fail-stop \\
\midrule
\textbf{\bench{}}                        & \cmark & Product & \cmark & \cmark & DAG        & Graded & Tolerance \\
\bottomrule
\end{tabular*}
\end{table}

\section{Related Work}
\label{sec:related}

\textbf{Coding agents and long-horizon software engineering.}
Evaluation of coding models has progressed from isolated code generation to agentic
software engineering, where models modify repositories, execute programs, and iteratively
refine their solutions. SWE-bench~\citep{jimenez2024swebench} evaluates repository-level
issue resolution, while more recent benchmarks extend the horizon of the work itself:
SWE-EVO~\citep{thai2025sweevo} specifies an entire software release, and
SWE-Marathon~\citep{desai2026swemarathon} studies ultra-long-horizon software tasks.
Other work explicitly evaluates sequences of interactions over persistent state.
EvoCode-Bench~\citep{evocodebench2026} maintains one workspace and agent session across
5--15 requests, SWE-Together~\citep{wu2026swetogether} reconstructs multi-request
development sequences from recorded user--agent interactions, and
StaminaBench~\citep{sobal2026staminabenchstresstestingcodingagents} and
SlopCodeBench~\citep{2026arXiv260324755O} study degradation over increasingly long
interaction horizons. These works establish the importance of evaluating coding agents
beyond individual tasks. \bench studies this problem in web development, where agents must
repeatedly extend the application they themselves have built.

\smallskip
\textbf{Web-application generation and evolving requirements.}
Web-generation benchmarks evaluate complete applications through observable behavior.
WebGen-Bench~\citep{lu2025webgenbench} pairs website-generation requests with GUI tests
executed against live applications, with subsequent work providing finer-grained functional
evaluation~\citep{liu2026webcoderbench,tran2026vibecodebench}.
Web-Bench~\citep{xu2025webbench} also considers sequentially dependent web-development
tasks, but its tasks are formulated primarily as developer-facing specifications that often
prescribe implementation details (example in App.~\ref{app:webbench-example}). In contrast, \bench requests are written at the user level: they fix the expected behavior, including seed data where tests need exact values, but never prescribe the implementation (files, components, class names, or layout), and their features are grounded in real-world websites.
Other benchmarks introduce iteration primarily as refinement toward an existing intent:
WebGen-Agent~\citep{lu2025webgenagent} repeatedly improves an application against a fixed
request; Asuka-Bench~\citep{wang2026asukabench} progressively exposes a predefined product
specification through feedback; and ICAE-Bench~\citep{peng2026icaebench} reveals
requirements drawn from an existing repository.
In \bench, each request may introduce new functionality and is applied to the codebase
produced by all preceding requests. Dependencies among requests are represented as a DAG,
where an edge $i\!\rightarrow\!j$ indicates that functionality requested at $j$ relies on
functionality introduced at $i$. Unlike DAGs used to organize evaluation criteria, such as
in Asuka-Bench~\citep{wang2026asukabench}, our DAG captures dependencies in the
application's evolution, is hidden from the evaluated agent, and supports tests that directly
exercise cross-request integration. Thus, each model must continually extend its accumulated
implementation while integrating new functionality with what it built earlier.

\smallskip
\textbf{Evaluating sequential development and user tolerance.}
Sequential benchmarks must decide what happens after an intermediate failure.
Web-Bench~\citep{xu2025webbench} and EvoCode-Bench~\citep{evocodebench2026}, for example,
terminate a trajectory after a failed verifier, while independently scoring every request
implicitly assumes that the user always continues. These correspond to extreme user models:
one abandons the application when a request falls below a fixed pass criterion, while the
other tolerates any failure. \bench instead scores each request from browser-observed
behavior on a graded scale, allowing the stopping rule itself to vary. For a tolerance
threshold $t$, the user continues while request quality remains above $t$ and leaves
otherwise. Because the tolerance of the user population is unknown, we run all requests
ungated, replay all thresholds offline, and integrate performance over $t$. Threshold
integration also appears in selective
prediction~\citep{geifman2017selective,traub2024augrc,zhou2025aurc}
and budget-constrained agent evaluation via the area under performance-versus-budget curves~\citep{cai2025agentbalance}, while multi-turn safety evaluation tracks survival over turns at a fixed failure threshold~\citep{hu2026safersteer}. These works sweep a property of the system, such as an abstention threshold or compute budget, or fix the threshold, whereas we sweep a property of
the \emph{user}: the minimum quality they accept before abandoning the application.
Our Grade and Survival scores therefore summarize long-horizon performance across user tolerances without
committing to an arbitrary stopping rule. Table~\ref{tab:benchmark-comparison} summarizes the comparison.

\section{\bench{} Benchmark}
\label{sec:method}

We design \bench around three desiderata that we argue are essential for evaluating contemporary web-development agents. \textbf{Realistic}: applications are grounded in real-world use cases, while follow-up requests are derived from functionality observed in real products, reflecting how users iteratively develop web applications through high-level natural-language requests. \textbf{Cumulative}: requests form long-horizon, interdependent sequences in which each new feature must be integrated with an evolving application while preserving previously implemented functionality. This setting tests whether agents can maintain correctness and coherence as the codebase grows, rather than solving each request in isolation. \textbf{Agentic}: each request is addressed through a contemporary coding-agent scaffold rather than a single model invocation, allowing the model to interact with the repository, execution environment, and other available tools over multiple steps, as is characteristic of modern coding-agent deployments.
In what follows, we describe data curation (Sec.~\ref{sec:data-curation}), the harness (Sec.~\ref{sec:harness}), the functional judge (Sec.~\ref{sec:eval}), and our metrics (Sec.~\ref{sec:metrics}).

\subsection{Data Curation}
\label{sec:data-curation}

\textbf{Grounding in real web-apps.}
We randomly subsample 40 of the 101 initial application specifications in WebGen-Bench~\citep{lu2025webgenbench}. Following automated verification and repair, followed by human curation, we retain 32 application trajectories (320 requests) spanning 16 of its 20 domains. We treat each selected specification as a \emph{seed}: an initial application description that we extend into a sequence of ten requests. To ensure that these requests correspond to capabilities found in real applications, we ground their design in deployed web-apps. For each seed application, a retrieval agent (Sonnet 4.6 with web search and page fetching tools) identifies two existing websites with a similar purpose and extracts three to four notable features from each that extend beyond the seed specification. A generator model (Fable 5) then adapts these features into follow-up requests for the benchmark. This grounding makes the resulting requests more representative of how real products evolve and, importantly, prevents different applications from collapsing to the same small set of generic additions. We describe the application selection and domain distribution in App.~\ref{app:domains}, the retrieval procedure in App.~\ref{app:grounding}, and study the effect of grounding in App.~\ref{app:realism}.

\begin{figure}[t]
\centering
\includegraphics[width=1.0\linewidth,trim={0 16bp 0 12bp},clip]{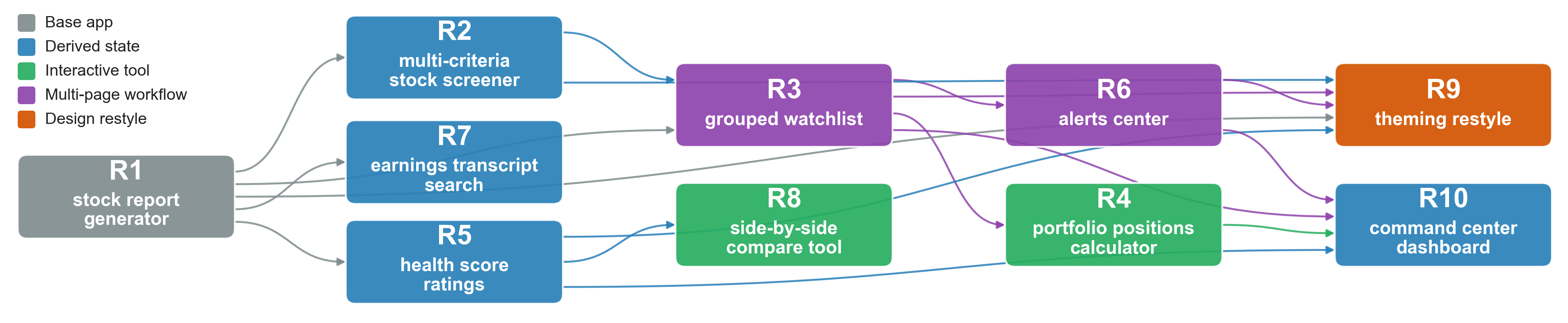}
\caption{\textbf{Requests within a flow form a dependency DAG.} Request 1 builds the base app and requests 2--10 add grounded features; an edge means the later feature cannot work unless the earlier one was built correctly. Colours mark the feature archetypes (App.~\ref{app:gen}).}
\label{fig:dag-example}
\vspace{-4pt}
\end{figure}

\smallskip
\textbf{Dependencies between requests.}
Grounding determines which features should be added to each application; we next determine how those features should depend on one another as the application evolves. In real applications, later functionality often relies on state, interfaces, or behavior introduced earlier. We therefore explicitly structure each request sequence as a directed acyclic graph (DAG) of feature dependencies. Before generating the natural-language requests, the generator constructs a DAG whose nodes correspond to the ten requested features and whose edges encode functional dependencies: an edge $i \rightarrow j$ indicates that request $j$ cannot be implemented correctly without functionality introduced by request $i$. We require these dependencies to be substantive rather than superficial; for example, merely operating on the same underlying data does not constitute an edge. To ensure that each sequence contains meaningful long-range dependencies, we require the DAG to contain a dependency chain of depth at least three and at least three later requests that depend on functionality introduced earlier. We validate the resulting DAG with a separate LLM-based verifier that is given the proposed feature dependencies and the constraints above. Sequences that fail verification undergo one automated revision and are discarded if they remain invalid. Human curators subsequently review all retained sequences and finalize their dependency structure and verification tests. The DAG is used only during benchmark construction and analysis and is never exposed to the evaluated agent. This design allows \bench{} to test whether agents can extend an evolving application while correctly reusing and preserving functionality introduced in earlier requests. We study these dependencies in Sec.~\ref{sec:abl-dag}; App.~\ref{app:gen} specifies the generation and verification procedure.

\smallskip
\textbf{Verification of incremental functionality.}
The sequential and dependent structure of \bench{} also changes how each request must be evaluated. We build on the functional evaluation protocol of WebGen-Bench~\citep{lu2025webgenbench}, which represents each test case as a UI operation paired with an expected outcome and uses WebVoyager~\citep{he2024webvoyager} to execute the test. WebGen-Bench constructs such tests to cover the requirements of a single, from-scratch application specification; our setting instead requires evaluating a feature added to an already evolving application. We therefore design five verification tests for each follow-up request: three test the newly introduced \textbf{functionality}, one tests its \textbf{integration} with an earlier feature along a dependency in the DAG, and one \textbf{backward} test re-runs a test from an earlier request. Because it scores earlier functionality in the current application, a failure may reflect either a regression or functionality that was never implemented, and a pass may follow a later repair; with one such test per request, it samples earlier functionality rather than exhaustively auditing regressions (verdict transitions in App.~\ref{app:backward-transitions}). For the 6\% of requests with no incoming dependency, the integration test instead checks a global rule introduced earlier (App.~\ref{app:gen}). To isolate the behavior under evaluation, we place any prerequisite application state directly in the user request as \emph{seed data}, rather than asking the evaluator to construct it through the UI. This keeps each test focused on the requested functionality and avoids conflating failures in test setup with failures of the feature itself. All tests are designed to be short and implementation-agnostic.

Together, these design choices yield a benchmark of grounded, interdependent request sequences with verification tailored to incremental development. Figure~\ref{fig:pipeline} summarizes the construction pipeline, while Table~\ref{tab:datastats} reports the resulting dataset composition and dependency statistics.

\vspace{-3pt}
\subsection{Agentic Harness}
\label{sec:harness}
\vspace{-2pt}
Prior web-generation benchmarks primarily evaluate applications generated from a single user request. In WebGen-Bench~\citep{lu2025webgenbench}, its best-performing Bolt.diy configuration follows a two-stage generation procedure: the model first selects a starter template and then generates the application from it. In \bench{}, each request is instead implemented through a multi-turn agentic interaction. The model can repeatedly inspect and modify the repository, execute commands, observe their outputs, and revise its implementation before completing the request. The repository and the conversation history persist across requests, so each interaction operates on the application produced by all preceding requests (execution details in App.~\ref{app:exec}). The agent receives no feedback on test outcomes between requests, so each request also measures how well it maintains what it has already built. Importantly, the first request in every flow is the original WebGen-Bench instruction, allowing us to directly examine the effect of the interaction protocol before any cross-request dependencies arise. On the shared initial requests, Haiku 4.5 achieves substantially higher functional accuracy with our multi-turn agentic harness than with the WebGen-Bench configuration. This observation motivates our use of an interaction protocol that allows models to inspect, execute, and iteratively refine their implementations.

\subsection{Request Evaluation}
\label{sec:eval}
After each request, we evaluate whether the resulting application satisfies it. We follow the evaluation protocol of WebGen-Bench~\citep{lu2025webgenbench}: a WebVoyager agent~\citep{he2024webvoyager} observes the interface through a Set-of-Marks representation, executes each test in the browser through Selenium, and returns YES, PARTIAL, or NO depending on whether the expected outcome is observed. We run it with a stronger computer-use backbone, Sonnet 4.6, which is among the strongest models on OSWorld-Verified (on par with Opus 4.6;~\citealp{xie2024osworld,anthropic2026sonnet46}) at a lower cost. WebGen-Bench validated this protocol against human annotators, reporting 86--94\% agreement, and we report a blinded human audit of the evaluator on \bench{} in App.~\ref{app:human-val}.

\subsection{Metrics}
\label{sec:metrics}

Each request $r \in \{1,\dots,R\}$ has $n_r$ tests, and test $k$ receives a score $s_{r,k}\in\{0,\tfrac{1}{2},1\}$ for fail, partial, and pass. The \emph{accuracy} of the request is its mean test score, $a_r=\frac{1}{n_r}\sum_{k=1}^{n_r} s_{r,k}$.

Averaging $a_r$ over requests ignores that development is sequential: after a poor result, a user may stop, and later requests are never reached. We model this with a tolerance threshold $t\in[0,1]$: the user continues only while every request so far scores at least $t$. The \emph{gate score} $g_r(t)$ indicates whether request $r$ is reached, \emph{delivered functionality} $Q(t)$ averages the accuracy of the reached requests, with unreached ones contributing zero, and \emph{request survival} $S(t)$ counts the consecutive requests, from the first, that score at least $t$:
\begin{equation}
g_r(t)=\mathbf{1}\Big[\min_{j<r} a_j \ge t\Big], \qquad Q(t)=\frac{1}{R}\sum_{r=1}^{R} g_r(t)\, a_r, \qquad S(t)=\sum_{r=1}^{R} g_r(t)\,\mathbf{1}[a_r\ge t],
\end{equation}
with $g_1(t)=1$. Since no single threshold fits every user, we integrate each curve over all thresholds to obtain its area under the curve (AUC) and use Grade as our primary metric; uniform integration weights all thresholds equally (App.~\ref{app:simple-metrics} gives closed forms and bounds under other weightings):
\begin{equation}
\mathrm{Grade}=\int_0^1 Q(t)\,dt, \qquad \mathrm{Survival}=\int_0^1 S(t)\,dt.
\end{equation}

\section{Experiments}
\label{sec:experiments}

\textbf{Experimental setup.}\label{sec:setup}
We evaluate 14 contemporary models on the 320 requests of \bench. Each model is deployed
with the OpenHands scaffold and is given a budget
of at most 20 agent turns for each request. The repository state is
preserved across the ten requests in a trajectory, such that every
request is implemented on top of the application produced by the
preceding interactions. After each request, we evaluate the resulting
application in the browser using WebVoyager with Sonnet~4.6 as the
underlying judge, following the verification protocol in Sec.~\ref{sec:eval} and Sec.~\ref{sec:metrics}.
We keep the benchmark instances, scaffold, interaction budget, and
functional evaluator fixed across models.

\subsection{Main results}
\label{sec:main-results-alt}

\begin{table}[t]
\centering
\caption{Model performance on \bench{}. The two rightmost columns give the percentage of requests whose build fails or whose served page raises a runtime error (Sec.~\ref{sec:process-errors}).}
\label{tab:main}

\scriptsize
\setlength{\tabcolsep}{4pt}
\renewcommand{\arraystretch}{0.95}

\begin{tabular*}{\linewidth}{@{\extracolsep{\fill}}lcccc|cc@{}}
\toprule
Model & Grade $\uparrow$ & Survival $\uparrow$ & Turns / request $\downarrow$ & \$ / sample $\downarrow$ & Build Err. \% $\downarrow$ & Runtime Err. \% $\downarrow$ \\
\midrule

Fable 5             & $\mathbf{0.694}$ & $\mathbf{7.32}$ & $\mathbf{4.7}$  & $14.40$ & $\mathbf{0.0}$ & $\mathbf{0.5}$ \\
Kimi K3             & $0.684$          & $7.31$          & $7.6$  & $4.17$  & $0.9$ & $3.6$ \\
GPT 5.6 Sol         & $0.622$          & $6.72$          & $9.0$  & $3.18$  & $0.8$ & $1.3$ \\
Opus 4.8            & $0.608$          & $6.50$          & $10.0$ & $8.85$  & $0.1$ & $3.0$ \\
Sonnet 4.6          & $0.607$          & $6.55$          & $10.1$ & $9.93$  & $0.4$ & $3.5$ \\
GLM 5.2             & $0.552$          & $5.93$          & $12.1$ & $4.02$  & $0.3$ & $5.4$ \\
DeepSeek V4-Pro     & $0.529$          & $5.84$          & $9.7$  & $1.66$  & $0.6$ & $3.9$ \\
DeepSeek V4-Flash   & $0.512$          & $5.80$          & $9.2$  & $0.12$  & $1.2$ & $6.1$ \\
Qwen3.8 2.4T-A95B   & $0.503$          & $5.42$          & $13.1$ & $5.93$  & $3.1$ & $15.4$ \\
GPT 5.6 Terra       & $0.488$          & $5.34$          & $6.5$  & $2.11$  & $0.3$ & $1.2$ \\
Haiku 4.5           & $0.377$          & $4.43$          & $13.1$ & $2.31$  & $1.6$ & $7.4$ \\
Qwen3.5 397B-A17B        & $0.315$          & $3.63$          & $6.5$  & $2.08$  & $0.9$ & $13.4$ \\
GPT 5.6 Luna        & $0.277$          & $3.12$          & $8.1$  & $\mathbf{0.36}$  & $0.6$ & $5.4$ \\
Qwen3.5 35B-A3B     & $0.162$          & $2.09$          & $9.4$  & $1.54$  & $2.5$ & $36.6$ \\

\bottomrule
\end{tabular*}
\end{table}

\textbf{Overall performance.}
On Grade, our primary metric, Fable 5 ($0.694$) and Kimi K3 ($0.684$) lead, followed by GPT 5.6 Sol ($0.622$), while the weakest reaches only $0.162$. We do not rank the top two, whose $0.010$ gap is within run-to-run variation (App.~\ref{app:variance}). Survival follows nearly the same order, with the best surviving about $7.3$ of $10$ requests (Table~\ref{tab:main}; standard deviations in App.~\ref{app:variance}). Although every test is fully passed in some run (App.~\ref{app:gen}), even the best models \textbf{do not sustain high-quality functionality across the trajectory}.

\begin{figure}[t]
\vspace{-2mm}
\centering
\includegraphics[width=0.93\linewidth]{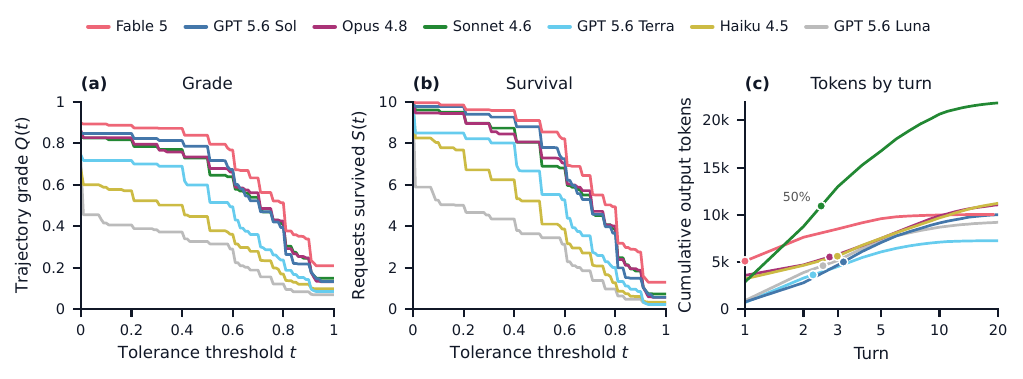}
\vspace{-1mm}
\caption{\textbf{Performance across tolerance levels and output length.} \textbf{(a, b)} Tolerance sweeps for the closed-source models (open-source models in App.~\ref{app:oss-sweeps}): delivered functionality $Q(t)$ and request survival $S(t)$ as a function of tolerance $t$; the area under each curve gives the Grade and Survival in Table~\ref{tab:main}. Curves show the run mean across three runs. \textbf{(c)} Cumulative output tokens per request by agent turn; dots mark where each model reaches half of its total.}
\label{fig:survival}
\vspace{-4pt}
\end{figure}

\smallskip
\textbf{Robustness to user tolerance.}
Figure~\ref{fig:survival} (left) sweeps the tolerance threshold $t$
for the closed-source models. Fable 5 leads at every $t$;
GPT 5.6 Sol, Opus 4.8, and Sonnet 4.6 stay close together throughout;
and Haiku 4.5 and GPT 5.6 Luna remain well below.
The gap between stronger and weaker models thus
persists across the full tolerance range, although their exact
ordering can change. App.~\ref{app:simple-metrics} reports mean accuracy ($t=0$) and fail-stop scores ($t=1$) for all models.

\textbf{Generation effort.}
Two of the three strongest models, Fable 5 and GPT 5.6 Sol, each produce only about 100K output tokens per trajectory, and Fable 5 also uses the fewest turns per request ($4.7$). Several weaker models spend more: Haiku 4.5 and Qwen3.8 take $13.1$ turns per request, and Sonnet 4.6 produces more than twice as many output tokens as Fable 5 (Figure~\ref{fig:survival}c). For every model, half of these tokens come in the first three turns, and later turns add progressively fewer. Higher scores thus come from \textbf{token and turn efficiency}, although Fable 5 costs the most per sample.

\textbf{Robustness to the judge.}
LLM judges are known to favor their own outputs~\citep{zheng2023judging,panickssery2024selfpreference,wataoka2024selfpreference}, which could bias \bench toward the grader's model family. We therefore re-evaluate all runs, replacing the Sonnet 4.6 judge of Sec.~\ref{sec:setup} with three judges from other vendors (App.~\ref{app:judge}). Absolute scores shift with the judge, but each alternative judge flips only 2-3 of the 28 model pairs, all among GPT 5.6 Sol, Opus 4.8, and Sonnet 4.6, which are within 0.015 Grade of each other. No judge favors its own family: Sonnet 4.6 ranks lower under its own judge than under two of the others. The ranking is thus \textbf{not an artifact of the judge's identity}.

\subsection{Effect of Request Language}
\label{sec:request-language}

\begin{wraptable}{r}{0.56\linewidth}
\vspace{-23pt}
\centering
\scriptsize
\setlength{\tabcolsep}{2.5pt}
\caption{Grade by request language. \textbf{First}, \underline{second}.}
\label{tab:request_language}
\begin{tabular}{@{}lccccc@{}}
\toprule
 & Sonnet 4.6 & Haiku 4.5 & GPT 5.6 Terra & GLM 5.2 & GPT 5.6 Luna \\
\midrule
English & \underline{0.607} & \underline{0.377} & 0.488 & 0.552 & \textbf{0.277} \\
\midrule
French  & 0.594 & \textbf{0.388} & \textbf{0.529} & 0.567 & 0.198 \\
German  & \textbf{0.618} & 0.330 & 0.427 & 0.445 & 0.262 \\
Chinese & 0.583 & 0.343 & \underline{0.509} & \underline{0.597} & \underline{0.268} \\
Hindi   & 0.604 & 0.297 & 0.508 & 0.524 & 0.221 \\
Hebrew  & 0.601 & 0.296 & 0.431 & \textbf{0.602} & 0.220 \\
\bottomrule
\end{tabular}
\end{wraptable}
Users often describe the functionality they want in their own language, even when the application itself is in English. We test whether this affects coding agents by translating the user requests of \bench into five other languages, while the application itself stays in English and the rest of the setup (Sec.~\ref{sec:setup}) is unchanged. Following \citet{feng2025tear}, Opus 5 translates each request sequence separately, then self-verifies and repairs the translation.

Table~\ref{tab:request_language} shows that only Sonnet 4.6 is insensitive to the request language, staying within $0.03$ of its English Grade in every language. Other models vary by language. GPT 5.6 Terra exceeds English in French, Chinese, and Hindi but loses up to $0.06$ in German and Hebrew; Haiku 4.5 loses up to $0.08$; and GLM 5.2 improves in Hebrew ($0.602$) but drops in German ($0.445$). Drops also occur in high-resource languages, such as GPT 5.6 Luna in French ($0.198$ vs.\ $0.277$), so \textbf{agents should be evaluated in the language their users will write in}.

\section{Analysis}
\label{sec:analysis}

\subsection{DAG-Structured Requests Analysis}
\label{sec:abl-dag}

\begin{table}[t]
\centering
\begin{minipage}[t]{0.47\linewidth}
\vspace{0pt}
\centering
\caption{\textbf{Grade by test type.}
Grade computed separately for each test type. bold =  lowest}
\label{tab:dag}
\scriptsize
\setlength{\tabcolsep}{4pt}
\begin{tabular}{lccc}
\toprule
Model & Backward & Functional & Integration \\
\midrule
Fable 5 & $0.706$ & $0.678$ & $\mathbf{0.630}$ \\
Kimi K3 & $0.692$ & $0.663$ & $\mathbf{0.632}$ \\
Opus 4.8 & $0.629$ & $0.598$ & $\mathbf{0.564}$ \\
GPT 5.6 Sol & $0.624$ & $0.596$ & $\mathbf{0.553}$ \\
Sonnet 4.6 & $0.597$ & $0.564$ & $\mathbf{0.524}$ \\
GLM 5.2 & $0.559$ & $0.533$ & $\mathbf{0.473}$ \\
GPT 5.6 Terra & $0.457$ & $0.431$ & $\mathbf{0.416}$ \\
Haiku 4.5 & $0.373$ & $0.335$ & $\mathbf{0.286}$ \\
\midrule
\textbf{Average} & $0.580$ & $0.550$ & $\mathbf{0.510}$ \\
\bottomrule
\end{tabular}
\end{minipage}\hfill
\begin{minipage}[t]{0.47\linewidth}
\vspace{0pt}
\centering
\scriptsize
\setlength{\tabcolsep}{3pt}
\caption{App-side failure modes (\% of each model's failed tests). A test counts as failed only if it scores 0. Top: \% of all tests that fail.}
\label{tab:failure-modes}
\setlength{\tabcolsep}{3pt}
\begin{tabular}{@{}lcccc@{}}
\toprule
 & Fable 5 & Opus 4.8 & Sonnet 4.6 & Haiku 4.5 \\
\midrule
Failed tests & 8.4 & 13.4 & 15.1 & 33.1 \\
\midrule
Missing data/UI & 23.6 & 21.6 & 21.3 & \textbf{31.2} \\
Incorrect data & 21.3 & 11.7 & 16.2 & 14.6 \\
Styling & 18.3 & 16.3 & 17.8 & 9.8 \\
Runtime/build err. & \textbf{2.2} & 17.3 & 17.0 & 13.7 \\
State sync & 8.5 & 7.3 & 8.1 & 12.0 \\
Unresp.\ control & 10.5 & 7.6 & 7.4 & 8.5 \\
Other & 15.5 & 18.2 & 12.2 & 10.1 \\
\bottomrule
\end{tabular}
\end{minipage}
\end{table}

\begin{figure}[t]
\centering
\includegraphics[width=0.95\linewidth]{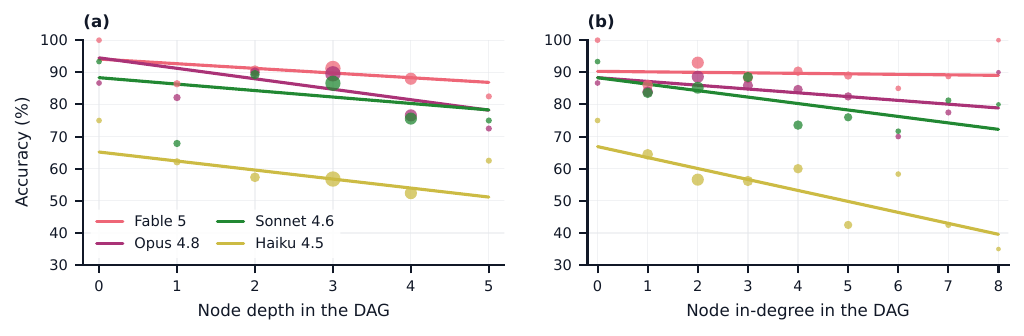}
\caption{\textbf{Accuracy by position in the dependency DAG.} Accuracy $a_r$ of requests 8--10 as a function of \textbf{(a)} the request's depth in the DAG and \textbf{(b)} its in-degree. Markers show the mean accuracy per value (area $\propto$ number of requests); lines are least-squares fits over all requests.}
\label{fig:dag-depth-indeg}
\end{figure}

Later features often consume the state, behavior, or interfaces of earlier ones, so the agent must compose new functionality with what it has already built. We ask whether this composition is harder than implementing a new feature or delivering earlier ones by comparing the three test types of Sec.~\ref{sec:data-curation}: functional, backward, and integration, the last of which traverses a DAG edge. We exclude the integration tests of requests with no incoming dependency (6\%), which check a global rule instead. For each type, we recompute Grade from that type's outcomes alone while keeping each request's gate score $g_r(t)$ based on its full accuracy $a_r$, so all three are compared over the same trajectories. Table~\ref{tab:dag} shows the same order for every model: backward tests score highest, followed by functional and then integration tests, with average Grade falling from $0.58$ to $0.55$ and $0.51$. Dependencies thus \textbf{add a composition challenge} beyond new or earlier functionality in isolation.

Since integration tests are the hardest, we next ask whether accuracy drops with dependency. We use two properties of each request's DAG node: its \emph{depth}, the longest dependency chain leading to it, and its \emph{in-degree}, the number of earlier requests it directly relies on. To separate dependency from position in the trajectory, we use only the last requests (8--10), which also cover the widest range of both. Figure~\ref{fig:dag-depth-indeg} plots their accuracy $a_r$ against each property, with a linear fit per model. Every line slopes downward. Together with Table~\ref{tab:dag}, this shows that difficulty compounds along the DAG: \textbf{requests that build on more earlier functionality are harder}, especially for weaker models.

\subsection{Build And Runtime Errors}
\label{sec:process-errors}

Functional tests show whether a feature works, not whether a failure comes from a broken application. We therefore also record, after each request, whether the production build fails (\emph{build error}) or the served application raises an uncaught exception, triggers a React error boundary, or fails to load a local module during browser evaluation (\emph{runtime error}). An error counts only for the requests after which it is present (e.g., a build error introduced at request 2 and fixed at request 3 counts for request 2 only); Table~\ref{tab:main} reports the percentage of the 320 request instances with each error type.

\begin{wrapfigure}{r}{0.47\columnwidth}
    \centering
    \vspace{-10pt}
    \includegraphics[width=\linewidth]{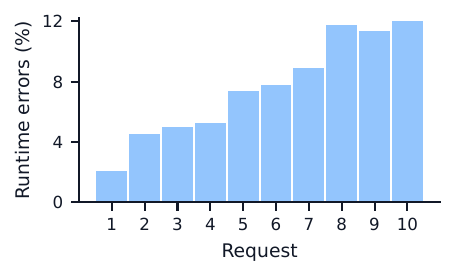}
    \label{fig:runtime_errors_per_request}
    \vspace{-30pt}
\end{wrapfigure}

Build errors are rare for most models, whereas runtime errors vary far more and track Grade more closely (Spearman $\rho=-0.72$ vs.\ $-0.52$ for build errors). Qwen3.5 35B-A3B, for example, has a build-error rate of $2.5\%$ but a runtime-error rate of $36.6\%$: models \textbf{more often preserve a buildable codebase than a reliably executable one}. Runtime errors also become more frequent as the application evolves, as the inset at right shows (runtime error rate by request index).

By contrast, \textbf{tool-call errors}, i.e., failed tool executions such as a string replacement that matches nothing, do not correlate with Grade (Spearman $\rho=-0.06$). The agent sees the failed call and repairs it in the next step, so these errors \textbf{cost steps rather than quality}.

\subsection{Failure Modes in Evolving Applications}
\label{sec:error-analysis-alt}

To understand why tests fail, we use Sonnet 5 to classify the failed tests (score 0) of Fable 5, Opus 4.8, Sonnet 4.6, and Haiku 4.5 into app-side failure modes (Table~\ref{tab:failure-modes}).
\emph{Missing data/UI}: a record, field, page, or control the test needs does not exist.
\emph{Incorrect data}: the data exists, but a value or computed result is wrong.
\emph{Styling}: a color, font, layout, or label differs from the request.
\emph{Runtime/build error}: the app or a route fails to build or crashes.
\emph{State sync}: an action is accepted, but the state or other views do not update.
\emph{Unresponsive control}: a rendered control does not react to input.

Haiku 4.5's failures are led by missing data ($31.2\%$, vs.\ $21.3$--$23.6\%$ for the other models), showing that the weakest model \textbf{struggles to integrate features across requests}: often, each page keeps its own copy of the data, so data added by one request never reaches pages added later. Fable 5, in contrast, rarely breaks the application ($2.2\%$ runtime or build errors, vs.\ $13.7$--$17.3\%$ for the other models), and its failures shift toward incorrect values and styling, where it has the highest shares of all four models. It thus makes the application work but \textbf{still fails on small details}.

\section{Conclusion}
\label{sec:conclusions}
We studied how reliably coding agents can extend persistent web applications over successive requests while integrating new functionality and preserving earlier behavior. \bench{} addresses this question through 320 cumulative requests across 32 applications grounded in real-world features and structured by functional dependencies. Browser-based tests assess each intermediate application, while tolerance-averaged metrics characterize trajectory performance without imposing a single stopping threshold.

Our experiments identify cross-request integration as a consistent challenge: agents perform worse on composing features across requests than on new or earlier-requested functionality alone. Later requests with deeper or more numerous dependencies are also solved less accurately, and runtime errors become more frequent as requests accumulate despite generally low build-error rates. These findings highlight feature integration and sustained runtime reliability as priorities for improving coding agents. 

Looking ahead, extending \bench{} to longer trajectories and closed-loop interaction, where users report failures and request repairs, would allow future work to examine whether agents can recover from mistakes while continuing to evolve an application.

\section*{Ethics Statement}
\bench{} does not involve human subjects. The benchmark builds on public WebGen-Bench specifications and on publicly visible features of existing websites, and all seeded data are fictional, so it contains no personal information. The generated applications run locally without external services. We describe our use of large language models below.

\section*{Reproducibility Statement}
The construction of \bench{} is described in Sec.~\ref{sec:data-curation}, with the grounding procedure in App.~\ref{app:grounding} and the full generation and verification specification in App.~\ref{app:gen}; the complete generator, verifier, and grounding prompts are included in the supplementary material. The agentic harness, the evaluation protocol, and the metrics are specified in Sec.~\ref{sec:harness}--\ref{sec:metrics}, and the experimental setup in Sec.~\ref{sec:experiments}. We report run-to-run variability in App.~\ref{app:variance}, human validation of the evaluator in App.~\ref{app:human-val}, robustness to the judge model in App.~\ref{app:judge}, and compute requirements in App.~\ref{app:infra}. The dataset, including all requests, verification tests, dependency graphs, and pinned seeds, as well as the code for running and scoring agents, are available at the links on the first page.


\section*{Acknowledgments}
We thank Roei Dalal for designing and building the parallel evaluation infrastructure that made the experiments in this paper fast to run.
\newpage

\bibliography{iclr2027_conference}
\bibliographystyle{iclr2027_conference}

\appendix

\definecolor{promptbg}{HTML}{F6F7F9}
\definecolor{promptrule}{HTML}{C5CCD6}
\definecolor{promptph}{HTML}{3D6A99}
\lstdefinestyle{prompt}{
  basicstyle=\ttfamily\scriptsize,
  backgroundcolor=\color{promptbg},
  frame=single, framerule=0.4pt, rulecolor=\color{promptrule},
  framesep=5pt, xleftmargin=5pt, xrightmargin=5pt,
  aboveskip=8pt, belowskip=8pt,
  breaklines=true, breakatwhitespace=true, breakindent=0pt,
  postbreak=\mbox{\textcolor{promptrule}{$\hookrightarrow$}\space},
  columns=fullflexible, keepspaces=true, showstringspaces=false,
  captionpos=t, abovecaptionskip=2pt, belowcaptionskip=2pt
}
\lstset{style=prompt}

\newpage
\section{Example request compared with Web-Bench}
\label{app:webbench-example}

Table~\ref{tab:webbench-example} contrasts a sequential task from Web-Bench~\citep{xu2025webbench} with a request from \bench{}. Both build on earlier work, but they differ in who the request addresses, where the feature comes from, how dependencies are defined, and how the result is checked.

\begin{table}[h]
\centering
\scriptsize
\caption{A Web-Bench task (React project, task 2) and an \bench{} request (link-in-bio site, request 6 of 10).}
\label{tab:webbench-example}
\renewcommand{\arraystretch}{1.25}
\begin{tabular}{@{}p{0.12\linewidth}p{0.40\linewidth}p{0.42\linewidth}@{}}
\toprule
 & \textbf{Web-Bench} & \textbf{\bench{}} \\
\midrule
Request &
\emph{``Create a appealing BlogList component that accepts an array of blogs as props and displays the titles in div elements with the className `list-item'. In Main.tsx, mock the blog data and display it using BlogList with this data: [\ldots]. Position BlogList on the left side of Main.tsx with a width of 300px.''} &
\emph{``I keep wanting my best stuff on top. Add a pin toggle on each link that floats it to the top of its group everywhere links are listed, plus Move Up and Move Down buttons that reorder links within their group---and the Preview should follow whatever order the manager shows. Unpinning a link should drop it back where it used to sit, not to the bottom, and nudging a link that's already at the top of its group shouldn't do anything weird. Pin Merch Store for me so it sits above Newsletter in Offers.''} \\
Written as & Developer specification: names files, components, class names, and layout sizes. & User request: describes behavior in product terms; no files, components, or selectors. \\
Feature source & Designed by the benchmark authors. & Adapted from real products (Linktree's pinned links, combined with Carrd's link reordering). \\
Dependency & Implicit: builds on the previous task's code. & Explicit DAG edges: pins and moves act within link groups (request~5), and the Preview mirrors the manager's order (request~3). \\
Tests & Playwright E2E scripts over fixed selectors. & Five browser-agent tests: three functional (e.g., unpinning returns a link to its former position, not the bottom), one integration (a move in the manager is reflected in the Preview), and one backward (a scheduled link's visibility from an earlier request). \\
Feedback & Failing tests are returned for one retry. & None between requests. \\
\bottomrule
\end{tabular}
\end{table}

\section{Application Selection and Domain Coverage}
\label{app:domains}

WebGen-Bench~\citep{lu2025webgenbench} provides 101 application
specifications across 20 domain classes, ranging from brochure sites
and storefronts to dashboards, forums, and browser games. We randomly
subsample 40 specifications and use each as a seed for constructing
a candidate trajectory of ten cumulative user requests, following
the procedure in Sec.~\ref{sec:data-curation}.

\paragraph{Verification and exclusions.}
Each candidate undergoes automated verification of its requests,
dependency graph, and verification tests against the construction
requirements. Candidates with identified violations receive one
repair pass and are verified again. A candidate is discarded if
unresolved violations remain, such as dependencies on functionality
not introduced by earlier requests, inconsistencies between seeded
data and expected test outcomes, or tests that require behavior
not specified in the requests. Human curators subsequently review
the remaining candidates for coherence and suitability, excluding
sequences whose follow-ups substantially duplicate existing
functionality or whose dependencies remain contrived rather than
reflecting plausible application development. Across these stages,
eight candidate trajectories are excluded, leaving 32 in the
final benchmark.

\begin{figure}[H]
\centering
\includegraphics[width=\linewidth]{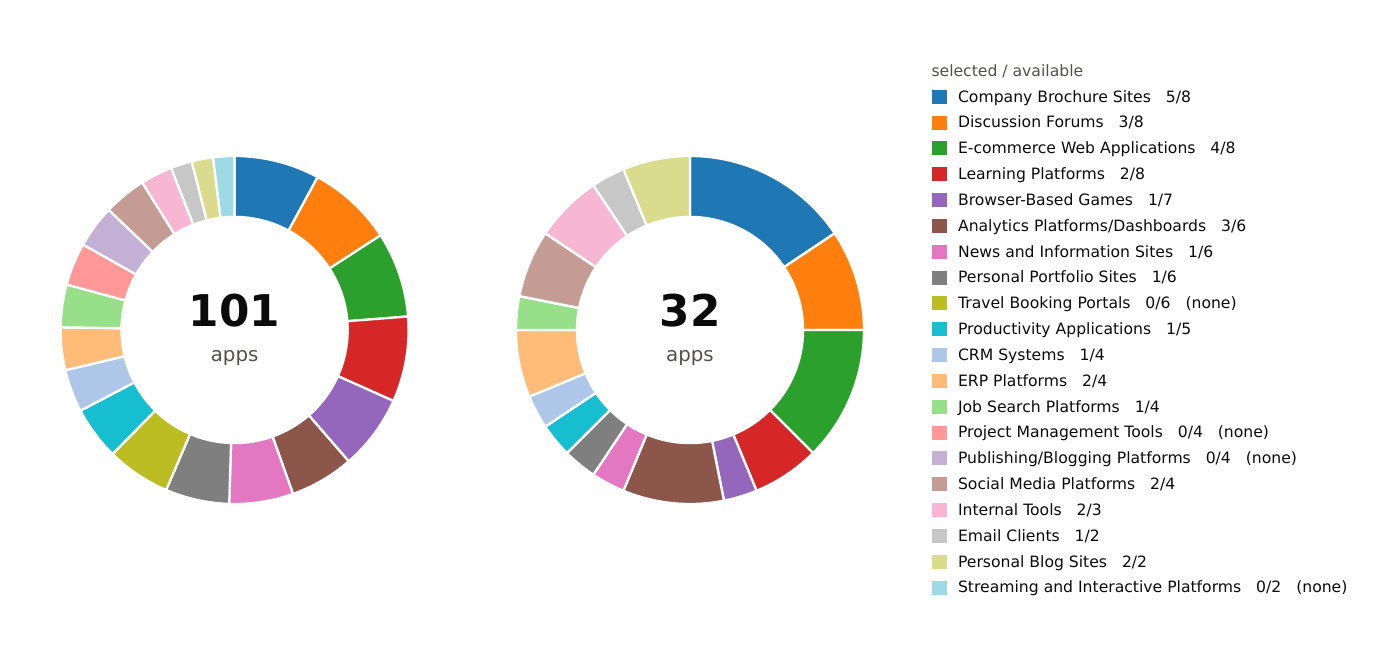}
\caption{\textbf{Domain coverage of the 32 flows.} Application-type distribution of our 32 flows (right) against all 101 WebGen-Bench specifications (left), sharing one colour per class. The legend gives selected over available counts per class.}
\label{fig:domains}
\end{figure}

\paragraph{Domain coverage.}
Figure~\ref{fig:domains} compares the domain distribution of the
32 retained trajectories with that of the full WebGen-Bench
collection. The final benchmark covers 16 of the 20 domain classes.
Under WebGen-Bench's broader three-way taxonomy, user-interaction,
content-presentation, and data-management applications account for
50\%, 25\%, and 25\% of our benchmark, respectively, compared with
approximately 49\%, 28\%, and 24\% of the source collection.

The represented classes, with the number of trajectories in each,
are Company Brochure Sites (5), E-commerce Web Applications (4),
Analytics Platforms and Dashboards (3), Discussion Forums (3),
ERP Platforms (2), Internal Tools (2), Learning Platforms (2),
Personal Blog Sites (2), Social Media Platforms (2),
Browser-Based Games (1), CRM Systems (1), Email Clients (1),
Job Search Platforms (1), News and Information Sites (1),
Personal Portfolio Sites (1), and Productivity Applications (1).
The four unrepresented classes are Travel Booking Portals,
Project Management Tools, Publishing and Blogging Platforms,
and Streaming and Interactive Platforms.

\section{Grounding: collecting similar websites and features}
\label{app:grounding}

Follow-up requests in \bench{} should reflect how real products actually grow, not how a language model imagines they might. We therefore ground each flow in features of existing websites that serve a similar purpose to the seed application. Grounding runs once per seed specification, before any request is written, and produces the real-product material that the generator (App.~\ref{app:gen}) later adapts into requests. It is a fixed, schema-constrained retrieval procedure executed by a language-model agent, Claude Sonnet 4.6, equipped with two tools: a web-search tool that returns ranked results for a text query, and a page-fetch tool that returns the content of a URL. The agent receives the seed's one-sentence summary, its full build prompt, and its WebGen-Bench application category, and proceeds in four steps.

\begin{enumerate}
\item \textbf{Query.} The agent issues web-search queries for deployed products that serve the same purpose as the seed. Similarity is judged by the agent against the seed's description and category, not by search rank, so the two sites it keeps need not be the first two results.
\item \textbf{Select.} From the results it selects two websites and for each, the product name, its URL, and a one-sentence description of what it does.
\item \textbf{Extract.} It fetches each selected site and identifies two to four notable features beyond the seed's core functionality (in practice, 57 of the 64 sites yielded four and 7 yielded three), keeping only features that are implementable in a single-page web app, that are progressive enhancements rather than core functionality, that a user would plausibly request after seeing the initial app, and that span a range of difficulty.
\item \textbf{Record.} For each feature it writes a name, a description of what the feature does, and a concrete follow-up idea phrased as a user request. The record is validated against a JSON schema that requires exactly two sites per seed and all three fields per feature.
\end{enumerate}

Over the 101 WebGen-Bench seeds this yields 202 sites and 801 features, with 195 sites contributing four features and the remaining seven contributing three. The 32 flows of \bench{} use the records of their seeds. The generator receives these records as its only source of feature ideas and never sees the retrieved pages. The agent's instructions are reproduced verbatim below.

\noindent\begin{minipage}{\linewidth}
\begin{lstlisting}[title={\small\textbf{Grounding prompt}}, moredelim={[s][\color{promptph}\itshape]{[}{]}}]
You are researching real-world websites similar to a web application concept. Your goal is to find 2 real, existing websites that are similar to this app and identify interesting features they have that could inspire follow-up feature requests.

APP: [short description]
Full description: [full description]
Category: [category]

Instructions:
1. Search the web for 2 real websites that serve a similar purpose to this app concept
2. For each website, visit it and identify 2-4 interesting/notable features
3. For each feature, think about how it could be turned into a realistic follow-up request in a multi-turn coding benchmark (e.g. "Add a dark mode toggle" or "Add CSV export for the data table")

Focus on features that are:
- Concrete and implementable in a single-page web app
- Progressive enhancements (not core functionality)
- Things a real user would ask for after seeing the initial app
- Varied in difficulty (some easy UI tweaks, some medium interactive features, some complex functionality)

Return the app index and its short description.
\end{lstlisting}
\end{minipage}

\subsection{Grounding ablation: realism and diversity}
\label{app:realism}

Grounding synthetic data in real-world sources is known to broaden its diversity~\citep{zhu2025real}. To test whether this holds for \bench, we compare it with an ungrounded counterpart built from the same 32 seed prompts and the same generator, but with the real-world features removed from the generator's prompt, keeping everything else fixed. All three judges \textbf{prefer the grounded requests} (69--74\%, with 95\% intervals above 50\%; Table~\ref{tab:realism}), and grounding \textbf{reduces the 4-gram overlap between unrelated apps by 29\%}.

\paragraph{Realism.}
Realism is a blind pairwise preference test. For each app, a judge sees the base build prompt followed by the grounded and the ungrounded sets of nine follow-up requests, labeled A and B, and picks the set whose features a mature real-world website in that category would have. The rubric defines such a feature and tells the judge to ignore writing style, length, and level of detail (prompt below). Because judges favor whichever set appears first, each pair is judged twice in each label order, giving $2 \times 2 \times 32 = 128$ judgments per judge. A judgment scores 1 when the grounded set wins, 0.5 for a tie, and 0 otherwise; scores are averaged per app, and the win rate is the mean over apps, with a 95\% bootstrap interval over apps. We use three judges from three different vendors, none of which is the author model's vendor. All three prefer the grounded requests, with intervals above 50\% (Table~\ref{tab:realism}).

\begin{table}[H]
\centering
\caption{Blind realism preference for the grounded requests, per judge, with a 95\% bootstrap interval over the 32 apps.}
\label{tab:realism}
\footnotesize
\setlength{\tabcolsep}{6pt}
\begin{tabular}{lcc}
\toprule
Judge & Grounded win rate (\%) & 95\% interval \\
\midrule
Grok 4.6       & 74.2 & [60.8, 86.7] \\
GPT 5.6 Sol    & 69.6 & [55.0, 82.5] \\
Gemini 3.1 Pro & 69.2 & [56.7, 80.8] \\
\bottomrule
\end{tabular}
\end{table}

\paragraph{Realism prompt.}
The system prompt is shared by every query and the user message is built per app.

\noindent\begin{minipage}{\linewidth}
\begin{lstlisting}[title={\small\textbf{Realism judge: system prompt}}]
You judge candidate follow-up requests for web applications. You see the original build prompt for an app, then two sets of follow-up requests a user might send afterwards. The sets come from two different generation methods; you are not told which is which and the order is randomised. Judge only the text shown, using this rubric.

What counts as a feature a real site would have:
- a capability you would expect on a mature, widely used product in this category, the common-sense core of that kind of site, plus the depth such a product accumulates
- it fits the domain and the audience the build prompt describes, consumer-facing vs internal admin
- it is something real users of that product would need, not a demo, toy or novelty mechanic

What counts as higher quality, when that is the question:
- implementable and internally consistent with the app described so far
- the requests build on each other rather than repeating or contradicting
- specific enough to act on, without inventing constraints a real user would not state

Explicitly NOT criteria, ignore them in every judgement:
- writing style, tone, register, or how much it sounds like a person typing
- length, level of detail, or the presence of exact values, seeded data and numbers
- the order the sets are shown in

Reply with JSON only: {"winner":"A"|"B"|"tie","confidence":1-5,"reason":"at most two sentences"}. Use "tie" only when the two sets are genuinely indistinguishable on the question asked.
\end{lstlisting}
\end{minipage}

\noindent\begin{minipage}{\linewidth}
\begin{lstlisting}[title={\small\textbf{Realism judge: user message (built per app)}}, moredelim={[s][\color{promptph}\itshape]{<}{>}}]
ORIGINAL BUILD PROMPT (turn 1):
<the app's base build prompt>

SET A - 9 follow-up requests:
1. <request>
...
9. <request>

SET B - 9 follow-up requests:
1. <request>
...
9. <request>

QUESTION: Which set asks for features you would actually find on a mature, real-world website in this category?

JSON only.
\end{lstlisting}
\end{minipage}

\paragraph{Diversity.}
We measure how much invented phrasing unrelated apps share. For app $a$, let $I_a$ be the sequence of 4-grams in its requests that do not appear in its base prompt, so phrasing inherited from the shared seed does not count as repetition. To compare equally sized sets, let $K=\min_a |I_a|$ and let $G_a$ be the set of the first $K$ 4-grams of $I_a$. With $N$ apps, the overlap is the average Jaccard similarity~\citep{jaccard1901} over app pairs,
\begin{equation}
\mathrm{Overlap} = \frac{2}{N(N-1)} \sum_{a<b} \frac{\lvert G_a \cap G_b \rvert}{\lvert G_a \cup G_b \rvert},
\end{equation}
in the spirit of Self-BLEU~\citep{zhu2018texygen}; lower means more diverse. Grounding reduces the overlap by 29\%.

\section{Dataset generation and verification}
\label{app:gen}

This appendix details how the follow-up requests and verification tests of Sec.~\ref{sec:data-curation} are generated and checked. Each request sequence is produced by an automated pipeline of generation, verification, and repair. A \emph{generator} agent, given only the base build prompt and the grounding features for the app, and never the built application, designs the grown app as a feature graph and writes the nine follow-up requests together with their verification tests, dependency graph, global rules, and pinned seed. A separate \emph{verifier} agent then re-derives compliance from the generated sequence alone, against a fixed rule set. If the verifier reports any violation, the generator makes one repair pass and the sequence is verified again; sequences that still fail are discarded. We summarize the specification that both agents follow below. The complete verbatim generator and verifier prompts are provided in the supplementary material and code release.

\begin{table}[h]
\centering
\caption{Dataset composition.}
\label{tab:datastats}
\begin{tabular}{lr}
\toprule
Property & Value \\
\midrule
Application build flows                 & 32 \\
User requests per flow                  & 10 \\
Verification tests per follow-up request & 5 \\
Extracted grounded features per app (mean) & 7.8 \\
Mean / max longest dependency chain (requests) & 4.75 / 6 \\
\bottomrule
\end{tabular}
\end{table}

\paragraph{Feature archetypes.}
The generator prompt requires every follow-up request to implement one of four archetypes, each with its own hardness condition: \emph{derived state}, a feature that computes, aggregates, or ranks over seeded or earlier records, with the seed arranged so that a naive implementation shows a visibly different result, \emph{interactive tool}, a feature driven purely by the user's typed input whose state updates during the interaction, \emph{multi-page workflow}, a flow that spans two or more views with every action reachable, and \emph{design restyle}, a bundle of four to six visual requirements applied consistently across at least three pages. The prompt asks for at least three archetypes per flow, at most four requests of any one, and no archetype twice in a row. Over the 288 follow-up requests of the 32 flows the split is 112 derived state (39\%), 77 multi-page workflow (27\%), 67 interactive tool (23\%), and 32 design restyle (11\%), exactly one per flow. Figure~\ref{fig:dag-example} colors the requests of one flow by archetype.

\subsection{Generator specification}
\label{app:gen-prompt}

\paragraph{Inputs.} The generator sees the base build prompt and the grounding entry of the app (similar websites and their features, App.~\ref{app:grounding}), but never the finished application, so it may not assume any implementation detail beyond what the build prompt requests. It favors grounded features that require real engineering, such as multi-criteria filtering, ranking with ties, aggregations, calculators, or pagination interacting with filters, and admits cosmetic features only within the design-restyle archetype.

\paragraph{Feature graph and dependencies.} The generator first imagines the app grown into a larger product and designs nine major features, each adapted, fused, or extended from a grounded feature, and then orders them topologically as requests 2--10. It records a dependency DAG over all ten requests, with request 1 as the base app and an edge for every \emph{significant} reliance: the later feature could not do what its request asks if the earlier feature were missing or broken. Merely reading the same seeded records does not count as a dependency. Edges point only forward and are labeled with what is relied on; the DAG is used for auditing and is never shown to the evaluated agent. Each sequence contains at least three dependencies between follow-up requests and at least one chain of depth at least three from the base app, and every design-restyle request has an edge to each feature it restyles. Dependencies must also be \emph{fair}: they run only through user-visible behavior that an earlier request explicitly asked for, so any valid implementation of earlier requests supports the later ones.

\paragraph{Intrinsic difficulty.} Every follow-up must be a distinct new capability, a different core operation on a different UI surface from the base app and from every other request, and must bundle several interacting requirements rather than a single control. Its correctness must also hinge on a non-trivial sub-problem that a shallow implementation gets wrong: boolean composition of criteria, counts that stay consistent with the visible set, stable sorting with ties, exact computed values with correct rounding and units, empty, boundary, or first-load handling, or precise input validation (for design restyles, full cross-page consistency). Depending on an earlier request does not by itself satisfy this requirement.

\paragraph{Pinned seed and global rules.} Exactly one early request (request 2 or 3) pins a small dataset of about five to eight records with exact names and values, so that later derived-state requests have exact expected results. Three app-wide \emph{global rules} per flow, such as a domain-specific formatting, labeling, or ordering convention, are introduced as user preferences in early requests with an explicitly app-wide scope. Later requests never restate them, so compliance is tested silently.

\paragraph{Feasibility and style.} Every feature must be implementable as a self-contained client-side React application with local seeded data: no external APIs or live data, LLM-based features, messaging, authentication, file upload or download, real-time synchronization, or dependence on the current date, and exports are rendered on screen. Requests are typically written in two to five sentences in a casual product-owner voice, refer to earlier work as a user would, and embed all exact values conversationally rather than as specification clauses.

\paragraph{Verification tests.} Each follow-up has exactly five tests in a fixed order. Tests~1--3 are \emph{functional} and exercise the new request; at least one of them targets its non-trivial sub-problem, and from the pinned-seed request onward, all but two of the benchmark's requests include an \emph{exact-value} test whose expected constant is computed from the pinned records, with the seed arranged so that a naive build shows a different constant. Test~4 is an \emph{integration} test that walks a DAG edge into the request, acting on the earlier feature and checking that its effect appears correctly in the new one; for requests with no incoming dependency, this slot instead checks a global rule on the new surface. Test~5 is a \emph{backward} test copied verbatim from an earlier request (for request 2, a test of the base app), with targets chosen to favor features that later requests rely on; at least five of requests 2--9 are re-tested later in the sequence. Tests must be executable by the click-and-type browser agent (no hover, drag, sliders, or keyboard navigation) within about six operations, use the simplest interaction that proves the check, and never ask the agent to set up state or invent values; any data a test needs is seeded through its request. At most two tests per sequence may check a list-count convention. Finally, tests are \emph{implementation-agnostic}: they reference only elements and values that the requests explicitly asked for, and every expected result must hold for any reasonable implementation of the requests.

\paragraph{Construction-time interaction audit.} During benchmark construction, to check the tests against the browser evaluator, a team of research engineers manually inspected approximately 200 randomly sampled browser-test executions spanning all test types (functional, integration, backward, and global-rule tests), reviewing interaction traces, visual evidence, and assigned scores for consistency with the intended checks. This qualitative audit informed the interaction constraints above: tests exclude hover-dependent behavior, drag-and-drop and drag-to-reorder interactions, keyboard navigation, and controls such as sliders, range handles, and dials; numeric thresholds are instead specified through text inputs, and tests requiring roles or hidden states inaccessible to the evaluator are excluded. These constraints apply to every test type and align the tests with the evaluator's supported click-and-type interactions. After these revisions, we conducted the quantitative human-validation study described in Appendix~\ref{app:human-val}.

\paragraph{Test satisfiability.} To check that the tests can be passed, we pool all runs of the five strongest models (Fable 5, Kimi K3, GPT 5.6 Sol, Opus 4.8, and Sonnet 4.6). Every verification test receives a full score in at least two of these runs. For 284 of the 288 follow-up requests, a single application passes all five of the request's tests at the same point in a trajectory, showing that the tests are jointly satisfiable. We manually inspected the remaining four requests and found no conflicting requirements: each of their tests is passed individually, though no single run passes all of them at once.

\paragraph{Backward-test transitions.}\label{app:backward-transitions} Each backward test re-runs a test that was already scored when its original request was evaluated. Pooling all runs of all models, we compare each backward test's verdict at that earlier evaluation with its verdict when re-run at a later request (Table~\ref{tab:backward-transitions}). Of the tests that passed earlier, 94.8\% still pass and 5.2\% no longer do (3.8\% NO, 1.5\% PARTIAL), the rate of apparent regressions. Of the tests that failed earlier, 29.0\% improve, 21.0\% to a full pass, reflecting repairs made while implementing later requests. A failed backward test therefore does not by itself indicate a newly introduced regression. Because both verdicts come from the automated judge, some transitions reflect evaluator variability rather than changes in the application (App.~\ref{app:human-val}).

\begin{table}[h]
\centering
\footnotesize
\caption{Backward-test verdict transitions, pooled over all runs of all models. Rows: verdict at the earlier evaluation; columns: verdict when re-run as a backward test. Percentages are conditional on the earlier verdict.}
\label{tab:backward-transitions}
\begin{tabular}{lccc}
\toprule
 & \multicolumn{3}{c}{Later} \\
\cmidrule(l){2-4}
Earlier & YES & PARTIAL & NO \\
\midrule
YES     & 94.8\% & 1.5\%  & 3.8\%  \\
PARTIAL & 21.3\% & 58.6\% & 20.1\% \\
NO      & 21.0\% & 8.1\%  & 71.0\% \\
\bottomrule
\end{tabular}
\end{table}

\subsection{Verifier checks}
\label{app:verify-prompt}

The verifier reads the base build prompt and the generated sequence and audits the sequence against twelve rules, reporting each violation with its request and test: (1)~no banned controls or interactions (sliders, drag, hover, keyboard navigation); (2)~nine follow-ups, each with five tests in the order three functional, one integration, one backward; (3)~backward tests copied character-for-character from the cited request, with at least five of requests 2--9 re-tested; (4)~a valid DAG over all ten requests with forward edges only, depth at least three, and at least three dependencies between follow-ups, where every edge is significant, fair, and consistent with the request text; (5)~a pinned seed, with every exact-value constant recomputed by the verifier from the seeded records; (6)~two to three domain-specific global rules, introduced early, never restated, and checked on new surfaces by at least two later requests; (7)~a casual user voice, the archetype-mix constraints, and minimal tests with at most about six operations and no setup or free choices; (8)~feasibility as an offline client-side application whose results are observable on screen; (9)~intrinsic difficulty and the archetype-specific conditions; (10)~distinctness from the base app and from every other request; (11)~implementation-agnostic tests; and (12)~sufficient per-request complexity, so that no request is widget-sized or a trivial variation of standard CRUD functionality. A sequence enters the benchmark only if it passes all twelve checks, after at most one repair pass.

\subsection{Human validation of the evaluator}
\label{app:human-val}

After the construction-stage revisions to the tests and interaction constraints (App.~\ref{app:gen-prompt}), we validated the browser evaluator against human judgments. We randomly selected 300 executions from the full evaluation pool, across all models, applications, and request positions, balanced across test types: 100 functional, 100 integration, and 100 backward tests. Integration examples exclude global-rule checks, matching Sec.~\ref{sec:abl-dag}. Several research engineers annotated the executions, and each execution received one final human label. Annotators saw the request, the specific test, and the browser agent's screenshots and recorded actions; the benchmark verdict was hidden. When the screenshots were insufficient, annotators inspected the application directly; this occurred for 39 executions (13\%), and their final labels are included in all results below. Human and benchmark scores use NO $=0$, PARTIAL $=0.5$, and YES $=1$.

\begin{table}[h]
\centering
\footnotesize
\caption{Human validation of the evaluator on 300 executions. Mean judge $-$ human is the signed difference in test score; positive values indicate more lenient benchmark scoring.}
\label{tab:human-val}
\setlength{\tabcolsep}{3.5pt}
\begin{tabular}{lcccccc}
\toprule
Type & $n$ & Exact agreement & $\kappa$ & Linear-weighted $\kappa$ & Mean judge $-$ human & 95\% CI \\
\midrule
Functional  & 100 & 87.0\% & 0.69 & 0.76 & $-0.005$ & $[-0.055, +0.045]$ \\
Integration & 100 & 88.0\% & 0.74 & 0.81 & $+0.050$ & $[+0.009, +0.091]$ \\
Backward    & 100 & 88.0\% & 0.71 & 0.76 & $+0.025$ & $[-0.027, +0.077]$ \\
\midrule
Pooled      & 300 & 87.7\% & 0.72 & 0.78 & $+0.023$ & $[-0.004, +0.051]$ \\
\bottomrule
\end{tabular}
\end{table}

\noindent\begin{minipage}[t]{0.63\linewidth}
\vspace{0pt}
Table~\ref{tab:human-val} reports agreement and signed scoring differences by test type. Positive signed differences indicate more lenient benchmark scoring. Confidence intervals are paired $t$-intervals over individual executions, using sample standard deviations and $n-1$ degrees of freedom; exact agreement and $\kappa$ are point estimates. The pooled row summarizes the balanced audit sample, and overall linear-weighted agreement is 92.0\%. Human mean scores are 0.775 for functional tests and 0.720 for integration tests, whereas benchmark means are 0.770 for both. Exact agreement is 95.0\% for the 261 executions assessed from screenshots and 38.5\% for the 39 requiring live inspection. The latter were selected because screenshot evidence was insufficient and therefore form an ambiguity-selected subset, not a random sample. Table~\ref{tab:human-confusion} gives the overall confusion matrix.
\end{minipage}\hfill
\begin{minipage}[t]{0.33\linewidth}
\vspace{0pt}
\centering
\footnotesize
\captionof{table}{Overall confusion matrix of the human validation (rows: human score; columns: benchmark score).}
\label{tab:human-confusion}
\begin{tabular}{lccc}
\toprule
& \multicolumn{3}{c}{Benchmark} \\
\cmidrule(l){2-4}
Human & 0 & 0.5 & 1 \\
\midrule
0   & 48 & 7  & 7   \\
0.5 & 5  & 11 & 10  \\
1   & 4  & 4  & 204 \\
\bottomrule
\end{tabular}
\end{minipage}

In the screenshots and recorded actions, annotators found no navigation errors among the 261 executions assessed from screenshots; navigation errors appeared only among the 39 executions that required live inspection, where disagreements also concentrate (24 of the 37).

\section{Execution details}
\label{app:exec}

We run all coding agents in OpenHands with its recommended configuration and a budget of at most 20 agent turns per request. The repository and the agent's conversation history persist across the ten requests of a trajectory. Context compaction is handled by OpenHands. Reasoning and sampling settings for each model follow its vendor's recommended settings. For evaluation, each browser test runs in a fresh browser session, so no browser state carries over between tests.

\section{Run-to-run variability}
\label{app:variance}

Coding agents are stochastic, so a single run may not reflect a model's typical performance. To measure this variance, we repeat the full evaluation three times for the closed-source models. Table~\ref{tab:variance} reports the mean and the population standard deviation across three independent runs for the Anthropic and GPT 5.6 models of Table~\ref{tab:main}, run through the official Anthropic and OpenAI APIs. The open-source models were run once. Fable 5 is the most stable model, with a Grade standard deviation of 0.004, while Sonnet 4.6 and Opus 4.8 vary by about 0.025 between runs, so the 0.001 gap between them in Table~\ref{tab:main} is within noise, whereas the gaps between tiers, of 0.05 and more, are about two standard deviations of the noisiest models and wider for the others. With only three runs per model, these standard deviations are themselves coarse estimates. They capture run-to-run variation only, not the choice of the 32 applications, and open-source models were run once, so Grade gaps of about 0.01 (e.g., Fable 5 vs.\ Kimi K3) do not establish a ranking. The language and DAG analyses likewise lack confidence intervals.

\begin{table}[H]
\centering
\caption{Mean and population standard deviation across three runs for the closed-source models of Table~\ref{tab:main}.}
\label{tab:variance}
\footnotesize
\setlength{\tabcolsep}{6pt}
\begin{tabular}{lcccc}
\toprule
Model & Grade & Survival & Turns / request & \$ / sample \\
\midrule
\multicolumn{5}{l}{\textit{Anthropic}} \\
Fable 5      & $0.694 \pm 0.004$ & $7.32 \pm 0.03$ & $4.7 \pm 0.1$  & $14.40 \pm 0.17$ \\
Opus 4.8     & $0.608 \pm 0.023$ & $6.50 \pm 0.23$ & $10.0 \pm 0.2$ & $8.85 \pm 1.14$ \\
Sonnet 4.6   & $0.607 \pm 0.028$ & $6.55 \pm 0.29$ & $10.1 \pm 0.2$ & $9.93 \pm 1.16$ \\
Haiku 4.5    & $0.377 \pm 0.014$ & $4.43 \pm 0.19$ & $13.1 \pm 0.5$ & $2.31 \pm 0.47$ \\
\midrule
\multicolumn{5}{l}{\textit{OpenAI}} \\
GPT 5.6 Sol   & $0.622 \pm 0.007$ & $6.72 \pm 0.08$ & $9.0 \pm 0.9$  & $3.18 \pm 0.03$ \\
GPT 5.6 Terra & $0.488 \pm 0.032$ & $5.34 \pm 0.31$ & $6.5 \pm 0.8$  & $2.11 \pm 0.12$ \\
GPT 5.6 Luna  & $0.277 \pm 0.043$ & $3.12 \pm 0.37$ & $8.1 \pm 0.8$ & $0.36 \pm 0.09$ \\
\bottomrule
\end{tabular}
\end{table}

\section{Tolerance sweeps for the open-source models}
\label{app:oss-sweeps}

\begin{figure}[H]
\centering
\includegraphics[width=\linewidth]{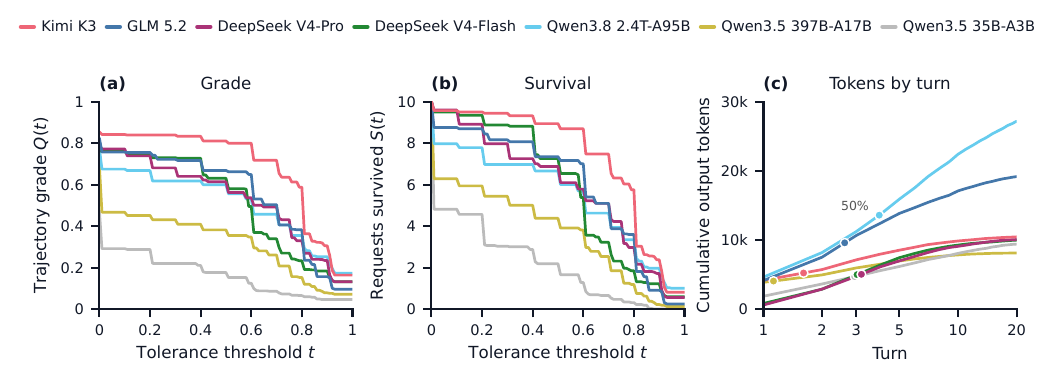}
\caption{\textbf{Tolerance sweeps for the open-source models.} Left: delivered functionality $Q(t)$ as a function of the tolerance threshold $t$. Right: request survival $S(t)$. The area under each curve is the Grade and Survival reported in Table~\ref{tab:main}. Single runs throughout.}
\label{fig:survival-oss}
\end{figure}

Grade and Survival (Sec.~\ref{sec:metrics}) integrate delivered functionality $Q(t)$ and request survival $S(t)$ over all tolerance thresholds $t$; plotting both against $t$ shows where along the tolerance range each model gains or loses. Figure~\ref{fig:survival} shows the tolerance sweeps for the closed-source models only, to keep the curves legible. Figure~\ref{fig:survival-oss} gives the same two panels for the seven open-source models of Table~\ref{tab:main}, six served on our own hardware and GLM 5.2 through its vendor's API. As in the main figure, the area under each curve is the model's Grade and Survival in the table.

\section{Sensitivity to User Tolerance}
\label{app:simple-metrics}

We examine how model rankings depend on user tolerance by evaluating
the two endpoints of the threshold range defined in
Sec.~\ref{sec:metrics}. At $t=0$, all requests contribute regardless
of earlier outcomes, so delivered functionality reduces to mean
request accuracy, $Q(0)=R^{-1}\sum_{r=1}^{R} a_r$.
At $t=1$, each request contributes only if all preceding requests
are fully successful. This yields the fail-stop score $Q(1)$,
which retains credit for the first imperfect request and assigns
zero to subsequent requests. Grade averages delivered functionality
over the full threshold range.

Table~\ref{tab:simple-metrics} reports these endpoint scores alongside
Grade. Mean accuracy and Grade produce similar rankings, with no
model moving more than two positions. The strictest threshold
produces larger changes: Qwen3.8 2.4T-A95B ranks ninth by Grade but
second by the fail-stop score, while Fable 5 ranks first under all
three metrics. These results show that individual model rankings
can depend substantially on assuming that users tolerate no
imperfect requests.

\paragraph{Threshold weighting and early errors.} Writing $m_r=\min_{j\le r}a_j$ with $m_0=1$, request $r$ is reached for $t\le m_{r-1}$ and survives for $t\le m_r$, so the two metrics have closed forms:
\begin{equation*}
\mathrm{Grade}=\frac{1}{R}\sum_{r=1}^{R}a_r\,m_{r-1}, \qquad \mathrm{Survival}=\sum_{r=1}^{R}m_r.
\end{equation*}
Uniform integration thus weights all thresholds equally; it is an aggregation convention rather than an assumption-free summary. Because $Q(t)$ is non-increasing in $t$, any other weighting of the thresholds yields a score between the fail-stop score $Q(1)$ and mean accuracy $Q(0)$, both reported in Table~\ref{tab:simple-metrics}. The closed form also makes the sensitivity to early requests explicit: the trajectories $(0.5,1,\ldots,1)$ and $(1,\ldots,1,0.5)$ both have mean accuracy $0.95$, but Grades of $0.50$ and $0.95$. This follows from the abandonment model, but it also means that an evaluator error on an early request affects more of the trajectory than one on a later request. Likewise, scoring PARTIAL as $\tfrac{1}{2}$ is a convention, and other values would shift absolute scores.

\begin{table}[t]
\centering
\caption{
Model performance at the endpoints of the tolerance range and
averaged across thresholds. Mean accuracy $Q(0)$ credits all
requests, whereas the fail-stop score $Q(1)$ excludes requests
after the first imperfect result. Models are ordered by Grade.
All scores lie in $[0,1]$; higher is better.
}
\label{tab:simple-metrics}
\begingroup
\small
\setlength{\tabcolsep}{10pt}
\renewcommand{\arraystretch}{1.15}
\begin{tabular}{lccc}
\toprule
Model
& \shortstack{Mean accuracy\\$Q(0)$}
& Grade
& \shortstack{Fail-stop score\\$Q(1)$} \\
\midrule
Fable 5
& \textbf{0.897} & \textbf{0.694} & \textbf{0.208} \\
Kimi K3
& 0.854 & 0.684 & 0.164 \\
GPT-5.6 Sol
& 0.848 & 0.622 & 0.133 \\
Opus 4.8
& 0.859 & 0.608 & 0.132 \\
Sonnet 4.6
& 0.844 & 0.607 & 0.148 \\
GLM 5.2
& 0.824 & 0.552 & 0.096 \\
DeepSeek V4-Pro
& 0.787 & 0.529 & 0.131 \\
DeepSeek V4-Flash
& 0.765 & 0.512 & 0.132 \\
Qwen3.8 2.4T-A95B
& 0.774 & 0.503 & 0.173 \\
GPT-5.6 Terra
& 0.743 & 0.488 & 0.084 \\
Haiku 4.5
& 0.673 & 0.377 & 0.098 \\
Qwen3.5 397B-A17B
& 0.689 & 0.315 & 0.071 \\
GPT-5.6 Luna
& 0.708 & 0.277 & 0.069 \\
Qwen3.5 35B-A3B
& 0.474 & 0.162 & 0.046 \\
\bottomrule
\end{tabular}
\endgroup
\end{table}
\section{Robustness to the judge model}
\label{app:judge}

Every result in this paper is graded by the same LLM-driven browser judge (Sec.~\ref{sec:eval}), so we test whether the results depend on which model runs it. LLM judges have been reported to prefer outputs from their own model or model family~\citep{zheng2023judging,liu2024narcissistic,panickssery2024selfpreference,wataoka2024selfpreference}. Judge choice also matters for web agents more broadly: on AgentRewardBench, no single LLM judge is best across benchmarks~\citep{lu2025agentrewardbench}. Since our default grader runs on Sonnet 4.6, which is also one of the evaluated agents, we re-evaluate the same runs with three judges from other vendors and compare the resulting Grades (Table~\ref{tab:judge}). Each alternative judge reverses the order of only 2-3 model pairs relative to the default, all among GPT 5.6 Sol, Opus 4.8, and Sonnet 4.6, which lie within 0.015 Grade of each other under the default judge. We also find no self-preference. Under its own judge, Sonnet 4.6 ranks fifth, below its third place under GPT 5.6 Terra and GLM 5.3 Flash, and GPT 5.6 Terra and GLM 5.2 hold the same rank under their own vendors' judges as under every other judge.

\begin{table}[h]
\centering
\caption{Grade of each agent under four judge models; Sonnet 4.6 is the default judge. \emph{Flips}: model pairs ordered differently than under the default judge.}
\label{tab:judge}
\footnotesize
\setlength{\tabcolsep}{5pt}
\begin{tabular}{lcccc}
\toprule
Agent & Sonnet 4.6 (default) & GPT 5.6 Terra & GLM 5.3 Flash & Qwen3.5 397B \\
\midrule
Fable 5            & 0.694 & 0.703 & 0.658 & 0.582 \\
Kimi K3            & 0.684 & 0.638 & 0.620 & 0.529 \\
GPT 5.6 Sol        & 0.622 & 0.578 & 0.575 & 0.461 \\
Opus 4.8           & 0.608 & 0.582 & 0.562 & 0.501 \\
Sonnet 4.6         & 0.607 & 0.616 & 0.593 & 0.489 \\
GLM 5.2            & 0.552 & 0.517 & 0.490 & 0.416 \\
GPT 5.6 Terra      & 0.488 & 0.446 & 0.433 & 0.357 \\
Haiku 4.5          & 0.377 & 0.364 & 0.334 & 0.320 \\
\midrule
Average            & 0.579 & 0.556 & 0.533 & 0.457 \\
Flips              & --   & 3 & 2 & 2 \\
\bottomrule
\end{tabular}
\end{table}

\section{Compute cost}
\label{app:infra}

Each model is evaluated on 320 requests (32 flows $\times$ 10 requests), each with up to 20 agent turns followed by five browser tests, i.e., about 1{,}600 browser evaluations per model, each a multi-step session of screenshot-and-act interactions. Browser verification is CPU-bound: run serially on one machine, a single model takes about 20 hours end to end. We execute the 32 flows concurrently, each in a dedicated container with six vCPUs. Our Kubernetes-based infrastructure scales to hundreds of vCPUs, providing each sample with an isolated headless browser environment. This distributed architecture enables us to evaluate multiple models in parallel in under two hours. We thank \textbf{Roei Dalal} for his invaluable DevOps contributions to this infrastructure.

\end{document}